\pdfoutput=1

\documentclass[11pt]{article}

\usepackage{EMNLP2023}

\usepackage{times}
\usepackage{latexsym}

\usepackage{amsmath}
\usepackage{enumitem}

\usepackage[T1]{fontenc}

\usepackage[utf8]{inputenc}

\usepackage{CJKutf8}
\usepackage{tipa}

\usepackage{microtype}

\usepackage{inconsolata}

\title{Machine Translation Evaluation Benchmark for Wu Chinese: \\ Workflow and Analysis}

\author{Jay Hongjian Yu*\thanks{*Equal contribution \\\\ This work is licensed under the Creative Commons BY-SA 4.0 International License. Visit \url{https://creativecommons.org/licenses/by-sa/4.0/} to view a copy of this license. For any use beyond those covered by this license, obtain permission by emailing \href{mailto:info@oldi.org}{info@oldi.org}. Copyright is held by the owner/author(s). Publication rights licensed to the EMNLP Endowment. \\} \\
  University of Washington \\
  \texttt{hjyu@uw.edu} \\\And
  Yiming Shi* \\
  East China Normal University \\
  \texttt{10214507003@stu.ecnu.edu.cn} \\\And
  Zherui Zhou* \\
  Shanghai Normal University \\
  \texttt{zheruizhou@outlook.com} \\\AND
  Christopher Haberland \\
  University of Washington \\
  \texttt{haberc@uw.edu} \\}

\begin{document}
\begin{CJK*}{UTF8}{gbsn}

{\makeatletter\acl@finalcopytrue\maketitle}

\begin{abstract}
Although the population of Wu speakers is the second largest among languages in China, it is a textually under-resourced language, creating significant challenges for building machine translation systems supporting Wu.
In this paper, we describe our Wu Chinese contribution to the FLORES+\footnote{https://github.com/openlanguagedata/flores/} dataset to serve as a training corpus and evaluation benchmark for machine translation models and we demonstrate its orthographic compatibility with existing Wu data.
Our contributions include: (1) an open-source, manually translated dataset, (2) full documentations on the process of dataset creation and validation experiments, (3) preliminary tools for Wu Chinese normalization and segmentation, and (4) benefits and limitations of our dataset, as well as implications for other under-resourced languages.
The project codes are stored on Github.\footnote{https://github.com/HongjianYu/FLORES-WU}
\end{abstract}

\section{Introduction}

Wu Chinese is a Sinitic language spoken in Shanghai, Zhejiang, parts of Jiangsu, Anhui, and Jiangxi of China.
It represents a complex and internally divergent dialect group \citep{Pan:91} with around 83 million speakers \citep{Eberhard:24}.
Despite having a robust population of speakers, Wu Chinese has been facing a sharp decline in daily usage due to the promotion of Standard Chinese.
Meanwhile, Wu Chinese lacks a widely-accepted writing system and is not commonly written by native speakers, which has relegated Wu to becoming under-resourced with respect to text data.
In this data-scarce context, machine translation of Wu is an extremely challenging task.

To assist in the development of language models in Wu Chinese, we construct a FLORES+ benchmark dataset for Wu machine translation models and conduct evaluations that validate its utility via a language identification task.
FLORES+ is an extension of the initial FLORES-101 project by \citet{nllb-22}, aiming at expanding the coverage to more languages worldwide.
FLORES features fully aligned data directly translated from English Wikimedia and is consequently ideal for multilingual translation systems.
Currently, the FLORES+ benchmark covers $3$ language varieties written in Hanzi: Mandarin Chinese (Standard Beijing), Mandarin Chinese (Taiwanese), and Yue Chinese (Hong Kong Cantonese).
The addition of Wu Chinese would facilitate machine translation across these similar varieties.

After translating and proofreading the new Wu Chinese dataset, we validated its consistency and generalizability with respect to existing Mandarin, Wu, and Yue Wikimedia resources.
We also devised measures to normalize and segment Wu Wikimedia data in order to enhance their fidelity and provide a standardized dataset upon which to conduct our evaluations.
Finally, we discuss the results of the experiments and suggest future work on Wu Chinese.

\section{Language Overview}

\subsection{Wu Sounds and Wu Writings}

Wu Chinese is mutually unintelligible with other Sinitic languages such as Mandarin and Yue (Cantonese), but shares a common set of \textit{Hanzi} (Chinese characters) with these varieties.
A significant feature of Wu Chinese that differentiates it from other Sinitic languages is its three-way VOT contrast in the syllable-initial position and its glottalization of "checked tones" \citep{Norman:88}, inherited from Middle Chinese.
A lot of Hanzi in Wu Chinese bear two ways of pronunciation: \textit{Wendu} ("文读", literary reading) and \textit{Baidu} ("白读", vernacular reading).
Wendu is the borrowing of pronunciation from Northern and Jianghuai Mandarin dialects; Baidu is the indigenous pronunciation derived from antecedent tones.
For instance, "学" reads [\textctc\textlhtlongy\ae] in Standard Mandarin; as for Wu, it reads [\texthth o\textglotstop] (Baidu) in "学堂" (the traditional word for "school") and [i\textturna\textglotstop] (Wendu) in "学校" (the modern word for "school" borrowed from Mandarin).
Although [\textctc\textlhtlongy\ae] and [i\textturna\textglotstop] appear to be unrelated, the sound change is in fact systematically induced according to the phonotactic constraints of Wu Chinese.
Wendu and Baidu can occur in one single word too, as "大学" ("university") reads [d\textturna\ \texthth o\textglotstop] where [d\textturna] is Wendu and [\texthth o\textglotstop] is Baidu.

With the above prerequisite knowledge, it is logical to believe that most syllabary utterances in Wu have had a Hanzi representation, since most Middle Chinese pronunciations have traceable documentations.
However, through an evolution of sounds and lexicon over a thousand years, the \textit{Benzi} ("本字", original character) of many sounds have been lost.
To recover missing graphemes in Wu Chinese writings, \citet{Zhao:56} proposed a guideline that for every Wu utterance:
\begin{enumerate}[nosep]
  \item If the Hanzi of the utterance is known, use that Hanzi;
  \item Otherwise, use a known Hanzi of the same pronunciation in the target Wu variety.
\end{enumerate}
This logic has constituted the overarching principle of modern Wu orthographies.
In the second case where we have multiple Hanzi candidates, we adhere to the following rules based on 简明吴方言词典 \citep{Min:86}:
\begin{enumerate}[nosep]
  \item Use the Hanzi that historically appears in Ming-Qing literature;
  \item When no historical usage is found, pick a Hanzi that best subscribes to the semantic meaning;
  \item Otherwise, choose the Hanzi that most frequently appears in vulgar texts.
\end{enumerate}

\subsection{Dialectal Variations}

In the past, the Suzhou dialect (a sub-dialect of Su-Hu-Jia within Northern Wu) has been the prestige form of Wu Chinese.
Beginning from the late 20\textsuperscript{th} century, the Shanghai dialect (spoken in the urban central area of Shanghai, also a sub-dialect of Su-Hu-Jia) has served as the lingua franca of the surrounding regions \citep{Chen:15} because of the city's population and economic importance.
There have also been recent attempts by the community to create "Standard Wu", which closely relates to Taizhou, Shanghai, and Ningbo dialects\footnote{https://wuu.m.wikipedia.org/wiki/标准吴语/}.
That being said, none of the Wu varieties is influential enough to profoundly alter the pronunciations or lexicons of others, considering the hegemonic impact of Standard Mandarin.

As a result of multiple factors, Wu dialects have developed vastly divergent readings of the same Hanzi.
This becomes especially problematic when no Hanzi is registered to an utterance, i.e. finding a Hanzi of the same pronunciation in the local dialect is necessary.
To illustrate how complex the spelling variations can become, below is an \textit{incomplete} list of the pronunciations and corresponding common spellings of the location/time preposition in $33$ Wu dialects transcribed by \citet{Qian:92}:
\begin{equation*}
  \text{"Prep. of loc./time"}
    \begin{cases}
      \text{[l\textschwa\textglotstop]} & \text{"勒", "了"} \\
      \text{[l\textturna\textglotstop]} & \text{"辣", "拉", "垃"} \\
      \text{[li\textglotstop]} & \text{"立"} \\
      \text{[le], [l\ae]} & \text{"来"} \\
      \text{[k\textepsilon]} & \text{"该"} \\
      \text{[\texttslig\textepsilon\textsuperscript{e}], [\textdzlig\textepsilon]} & \text{"在"}
    \end{cases}
\end{equation*}
Inevitably, adopting one spelling standard means discarding the rest of equivalently common spellings.
For instance, [l\textturna\textglotstop\ l\textturna\textglotstop] "辣辣" is the prevalent spelling of the location/time preposition (double syllable) in the Shanghai dialect, whereas [l\textschwa\textglotstop\ l\textturna\textglotstop] "勒拉" is commonly accepted in many other dialects and therefore more frequently encountered.
We shall discuss the Chongming dialect which is the variety used in our dataset, and its corresponding orthography in the next section.

\subsection{Relevant Resources}
\label{sec:rel}

Before proceeding, we want to outline the resources available for natural language processing tasks related to Wu Chinese.
The foundation of Wu Chinese studies was laid by 赵元任 (Yuen Ren Chao) with his 现代吴语的研究 \citep{Zhao:56}.
Works by later scholars include 当代吴语研究 \citep{Qian:92} and others.
There is 简明吴方言词典 \citep{Min:86}, a dictionary that covers most Wu dialects with a light emphasis on the Shanghai dialect lexicon, as well as 上海话大词典 \citep{Qian:08} specifically for the Shanghai dialect.
Thanks to community efforts, there is a Wu Wikipedia\footnote{https://wuu.wikipedia.org/wiki/}, a forum\footnote{https://wu-chinese.com/bbs/}, and several online dictionaries\footnote{http://wu-chinese.com/minidict/index.php}\footnote{https://www.wugniu.com/dict} made by 吴语协会 and 吴语学堂.

\section{Methodology}

The FLORES+ Wu dataset is directly translated from English into Wu Chinese.
The target Wu variety is the Chongming dialect.
The Chongming dialect (or more broadly the Shadi dialect) is a Wu dialect within the Su-Hu-Jia division, spoken in Chongming, Haimen and Qidong districts as well as in some areas of Zhangjiagang \citep{Zhang:09}.
Although Chongming belongs to the Manucipality of Shanghai, the dialect is distinctive from the urban variety on many aspects and is known for preserving many rare characteristics of Middle Chinese.
There is also a large dedicated Chongming dictionary \cite{Zhang:14}, which unfortunately was not accessible to us during our efforts.

While the Chongming dialect is not the most used or researched Wu dialect, it is a representative one of Northern Wu along the dialect continuum, because it spreads in between Shanghai and Suzhou, where the two prestige forms of Northern Wu are used.
Besides, the lexicon of the Chongming dialect has a significant overlap with other dialects for its preservation of archaic features common to most Wu dialects.
In contrast, the Shanghai dialect is less generalizable to other Wu dialects due to its integration with other Chinese languages.
However, the overlap between the Chongming dialect and Southern Wu dialects might be less prominent.
As a result, our dataset may be less effective for Southern Wu linguistic tasks.

Data were equally distributed and translated by 2 native speakers of the Chongming dialect who have earned or are pursuing a university degree in English.
Both translators grew up in Chongming; one went to Putuo, Shanghai for high school and college, and the other went to Fengxian, Shanghai for college.
They mainly speak Wu at home, but also speak it with peers on occasion.
They are exposed to the Shanghai urban dialect as well as other local varieties.
All translated data were checked by a third independent Wu speaker.

The translators worked collaboratively on the task.
They mainly used 简明吴方言词典 \citep{Min:86} and then 上海话大词典 \citep{Qian:08} (see \ref{sec:rel}) if they were unable to recall the parallel Hanzi that represent the utterances.
Despite the discrepancy between the Chongming and Shanghai lexicons, the dictionaries provided enough context to determine the appropriate orthography.
For example, both "乃末" and "乃么" point to the same word [ne m\textschwa\textglotstop]; as both dictionaries only list "乃末", it was easy to make the choice.
When the two dictionaries' orthographies diverged, 简明吴方言词典 was prioritized.
When there were phonetic distinctions between the Chongming and Shanghai dialect and the original character was indeterminable, we made sure that the selected Hanzi had aligned pronunciation.
Noticeably, we did not use "勿" [v\textschwa\textglotstop] but used "弗" [f\textschwa\textglotstop] for the word of negation.
We were also committed to maintaining a balanced language register, as the translated content is formal, though Wu Chinese is usually colloquial.
We referred to the broadcasting-style Wu Chinese found in Shanghai and Suzhou to achieve the desired register.
Beside the task of translation, the translators  dedicated time to review Wu dictionary entries and online fora to grasp the overall construct of the written Wu landscape.
During the proofreading process, when an alternative translation is suggested, the translator responsible for the line would ask for community guidance from the aforementioned fora.
Proper wordings were selected according to the intuitive preferences of Wu native speakers from the community.
In total, we have translated and verified the linguistic accuracy all $997$ sentences in the \texttt{dev} set.

\section{Data Samples}

This sections lists the first $5$ lines of translation along with their English counterparts. \\
\vspace{-0.2cm}
\begin{enumerate}[nosep]
  \item 斯坦福医学院个科学家勒礼拜一公布一种可以按种类划分细胞个新个诊断家生个发明：一种可以用标准喷墨打印机大量生产，差弗多小到只有一美分一只个可印芯片。 \\ On Monday, scientists from the Stanford University School of Medicine announced the invention of a new diagnostic tool that can sort cells by type: a tiny printable chip that can be manufactured using standard inkjet printers for possibly about one U.S. cent each.
  \item 首席研究者认为伊作兴可以勒低收入国家稍为早发现癌症、肺结核、艾滋病、疟疾个病人，伊拉个乳腺癌等疾病个治愈率是只富裕个国家个一半。 \\ Lead researchers say this may bring early detection of cancer, tuberculosis, HIV and malaria to patients in low-income countries, where the survival rates for illnesses such as breast cancer can be half those of richer countries.
  \item JAS 39C 鹰狮战斗机勒当地辰光差弗多早晨九点半（中央时间两点半）撞向飞机跑道爆炸，葛咾商业航班个机场关闭。 \\ The JAS 39C Gripen crashed onto a runway at around 9:30 am local time (0230 UTC) and exploded, closing the airport to commercial flights.
  \item 搿只飞行员拨认为是空军中队长迪罗克利特帕塔维（Dilokrit Pattavee）。 \\ The pilot was identified as Squadron Leader Dilokrit Pattavee.
  \item 当地个媒体报道，一只机场个救火车勒回应个辰光翻倒哉。 \\ Local media reports an airport fire vehicle rolled over while responding.
\end{enumerate}

\section{Validation}

\subsection{Task}

Language identification models have frequently been trained to corroborate translation datasets' correspondence with other texts in the target language. We train a language identification model on the FLORES+ Wu dataset to show its compatibility with other Wu Chinese text resources.
The goal of our experimental setup is to train a model to correctly identify the language of an input sentence from among Mandarin, Wu, and Yue when prompted a sentence written in Hanzi. 

Experiments are split into three parts.
In each part, we trained three binary classification models: Mandarin-Wu, Mandarin-Yue, and Wu-Yue on their respective datasets, and a three-way classification model on all designated data.
We recorded the performance of the models in terms of their accuracy on unseen data, collected separately by languages. 

We broke down the experiments into distinct trials that reflect noteworthy characteristics of the training and evaluation data.
In part \ref{tab:flores-flores}, we conducted a $9$:$1$ split on FLORES+ datasets to test their internal consistency.
In part \ref{tab:wiki-flores}, we trained the model on Wikimedia data and tested on FLORES+ to showcase the compatibility of FLORES+ in its common use cases.
In part \ref{tab:flores-wiki}, we reversed the training and testing data in part \ref{tab:wiki-flores} to explore the extent of generalizability of FLORES+ given that it consists of small but parallel data.

\subsection{Dataset Processing}

We adopted two data sources, Wikimedia\footnote{https://dumps.wikimedia.org/} and FLORES+.

We downloaded Wikimedia database XML dumps for Mandarin, Wu, and Yue.
Since Mandarin and Yue dumps are significantly larger than Wu, only a portion of the data was extracted.
After normalization, each dataset comprises $25,000$ lines of texts.

FLORES+ datasets in use include the existing two Mandarin \texttt{dev} sets and Yue \texttt{dev} set, as well as the newly built Wu \texttt{dev} set.
As a result, there are $1994$ lines of sentences in Mandarin, $997$ in Wu and $997$ in Yue.

\subsubsection{Normalization}

The dumps were preprocessed with a simple filter removing Latin characters and template symbols.
Because Mandarin and Yue Wikimedia were written in Traditional Chinese and Wu Wikimedia partially, we utilized OpenCC\footnote{https://github.com/BYVoid/OpenCC/} which supports character-level and phrase-level conversion from Traditional to Simplified Chinese.
OpenCC conversion was also applied to FLORES+ Taiwan Mandarin and Yue datasets.

For the lack of a standard orthography, Wu Wikimedia requires an additional step of normalization.
For some characters and words that are often spelled differently, we replaced the constituents by our standard forms.
However, the brute force method does not work for every character and word that need normalization. For example, "勒海" before normalization could be interpreted differently:
\begin{equation*}
  \text{勒海}
    \begin{cases}
      \text{勒嗨} & \text{"In there, over there"} \\
      \text{勒海(浪)} & \text{"In the sea"}
    \end{cases} 
\end{equation*}
The ambiguity of the language results in demands on more advanced normalization schemes, which are essential for language models to grasp semantic understandings.

\subsubsection{Segmentation}

Since Chinese languages do not depend on spaces to separate words, segmentation tools tailored to the respective languages are indispensable.
For Mandarin, we used \texttt{jieba}\footnote{https://github.com/fxsjy/jieba/}, a popular open-source segmentation library with a prefix dictionary and a HMM-based model with Viterbi algorithm for unknown words; for Cantonese, we used \texttt{cantoseg}\footnote{https://github.com/ayaka14732/cantoseg/} which is built from \texttt{jieba} and reads in a merged corpus from \texttt{jieba} and \texttt{PyCantonese} \citep{Lee:22}.

As for Wu, We decided on adding an auxiliary dictionary to \texttt{jieba} for frequent words and phrases in Wu Chinese that are not present in Mandarin.
We found this approach has also been used by \citet{Chen:23}.
The entries in the auxiliary dictionary have been collected from 简明吴方言词典 \citep{Min:86} for its wide coverage on Northern and Southern Wu dialects and 上海话大词典 \citep{Qian:08} for its rich lexicon.

\subsection{Model}

We use fastText\footnote{https://fasttext.cc/} text classification \citep{joulin:16} for all experiments.
FastText is a CPU-based library for efficient learning of word representations and sentence classification.
We tagged language labels to individual lines in every dataset and called the \texttt{supervised} command to train the models.
When testing, fastText takes a $k$ parameter and returns both precision and recall rates within the first $k$ predicted labels.
As only $2$ or $3$ distinct labels were present in each run, we used $k$=$1$ to compute the classification accuracy.

\subsection{Results}

\begin{table}
\centering
\begin{tabular}{lclclclclcl}
\hline
& \textbf{cmn} & \textbf{wuu} & \textbf{yue} & \textbf{total} \\
\hline
\textbf{cmn-wuu} & $0.995$ & $0.990$ & - & $0.993$ \\
\textbf{cmn-yue} & $1.000$ & - & $0.970$ & $0.990$ \\
\textbf{wuu-yue} & - & $0.990$ & $0.990$ & $0.990$ \\
\textbf{c-w-y} & $0.995$ & $0.990$ & $0.979$ & $0.990$ \\
\hline
\end{tabular}
\caption{FastText classification precision rates by languages ($k$=$1$), trained with and evaluated on FLORES+ \texttt{dev} sets, $9$:$1$ split. Rows represent in what languages the model is trained with; columns represent the language of the testing data. We use the ISO 639-3 codes for abbreviation: cmn (Mandarin), wuu (Wu), yue (Yue).}
\label{tab:flores-flores}
\end{table}

In part \ref{tab:flores-flores}, all four models demonstrate a high accuracy in classifying all languages.
This validates the internal consistency of FLORES+ datasets including the new Wu Chinese addition.

\begin{table}
\centering
\begin{tabular}{lclclclclcl}
\hline
& \textbf{cmn} & \textbf{wuu} & \textbf{yue} & \textbf{total} \\
\hline
\textbf{cmn-wuu} & $0.896$ & $0.999$ & - & $0.930$ \\
\textbf{cmn-yue} & $0.968$ & - & $0.987$ & $0.975$ \\
\textbf{wuu-yue} & - & $0.996$ & $0.986$ & $0.991$ \\
\textbf{c-w-y} & $0.868$ & $0.997$ & $0.971$ & $0.926$ \\
\hline
\end{tabular}
\caption{FastText classification precision rates by languages ($k$=$1$), evaluated on FLORES+ \texttt{dev} sets, trained with Wikimedia texts. Rows and columns are the same as Table \ref{tab:flores-flores}.}
\label{tab:wiki-flores}
\end{table}

In part \ref{tab:wiki-flores}, we can observe a total accuracy over $90\%$ for all four models.
However, a drop in accuracy for the Mandarin-Wu model on Mandarin texts indicates false positives of Mandarin texts mislabelled as Wu.
Alternatively speaking, training on Mandarin and Wu Wikimedia data allows the model to capture features of Wu and thus correctly label Wu data, but is less effective for recognizing Mandarin features.

\begin{table}
\centering
\begin{tabular}{lclclclclcl}
\hline
& \textbf{cmn} & \textbf{wuu} & \textbf{yue} & \textbf{total} \\
\hline
\textbf{cmn-wuu} & $0.944$ & $0.639$ & - & $0.792$ \\
\textbf{cmn-yue} & $0.949$ & - & $0.796$ & $0.872$ \\
\textbf{wuu-yue} & - & $0.815$ & $0.884$ & $0.849$ \\
\textbf{c-w-y} & $0.929$ & $0.615$ & $0.735$ & $0.759$ \\
\hline
\end{tabular}
\caption{FastText classification precision rates by languages ($k$=$1$), evaluated on Wikimedia texts, trained with FLORES+ \texttt{dev} sets. Rows and columns are the same as Table \ref{tab:flores-flores}.}
\label{tab:flores-wiki}
\end{table}

In part \ref{tab:flores-wiki}, due to insufficient training data, the models exhibit tendency to misclassify Wu and Yue data as Mandarin.
There is a more significant contraction in testing accuracy on Wu data than Yue.
Meanwhile, the accuracy of Mandarin-Yue and Wu-Yue models is maintained at a relatively stable level.

Some typical misclassifications are listed below. These input data are Mandarin but mislabeled as Wu, presented in the segmented format.
The corresponding Wu translations are provided as well.
"cmn" denotes Mandarin data (mislabeled as Wu) and "wuu" denotes Wu data (correctly labeled).

\begin{enumerate}[nosep]
  \item (cmn) 人类\, 的\, 手\, 比脚\, 短\, ，\, 指\, （\, 趾\, ）\, 骨\, 更\, 直\, 。 \\ (wuu) 人个\, 手比脚\, 短\, ，\, 趾骨\, 更\, 直\, 。
  \item (cmn) 看到\, 有人\, 愿意\, 支持\, 我\, ，\, 我\, 很\, 高兴\, 。 \\ (wuu) 我蛮\, 高兴\, 有人\, 愿意\, 支持\, 我个\, 。
  \item (cmn) 越来越\, 多\, 超市\, 开始\, 提供\, 更\, 多样化\, 的\, 即\, 食品\, ，\, 部分\, 超市\, 甚至\, 提供\, 微波炉\, 或\, 其他\, 设备\, 以\, 食物\, 加热\, 。 \\ (wuu) 超市\, 里个\, 现成\, 食品\, 种类\, 越来越\, 多\, 。\, 有眼\, 超市\, 甚至\, 提供\, 微波炉\, 或者\, 其他\, 方式\, 来\, 加热\, 食物\, 。
\end{enumerate}

From these cases we can observe many common words in the two languages. There are nuanced differences in phrasing order and sentence structures but the presented orders and structures are generally acceptable in both languages and only subject to personal habits of the translators. Despite lexicon similarity, the model also seems to have difficulty in recognizing Wu constituents due to the relatively weak performance of the segmentation tool, evident in "我个" (1, 2), "我蛮” (2).

Overall, the FLORES+ Wu dataset is consistent and capable of evaluating models trained with common data after appropriate normalization and segmentation.
However, its benchmarking quality might not be as good as the Mandarin and Yue dataset for several reasons.

The tokenization scheme could be further optimized with a better segmentation tool in use.
The current manually configured word list for Wu in the auxiliary dictionary of \texttt{jieba} is relatively small compared to the pre-built Mandarin dictionary in \texttt{jieba} and the independently maintained Yue dictionary by \citet{Lee:22}.
As some syntactic structures of Wu have not been recognized by \texttt{jieba}, the models are unable to learn accurate representations of the constituents.

Although the spelling standard used in FLORES+ Wu dataset is relatively generalizable to other dialects, it fails to take account of many expressions in Southern Wu dialects which are a part of the Wikimedia data.
Therefore, we suggest that the training and testing datasets should align in the range of dialects whenever possible.

Moreover, the tendency of Wu Chinese to be influenced by Mandarin poses problems, exemplified by our classifier mislabeling Wu data containing "了" as Mandarin, because this grammar particle is common in Mandarin but infrequent in older Wu data.

\section{Conclusion}

As for now, a contemporary, consistent, and organized corpus becomes crucial for high-quality Wu Chinese language models. Meanwhile, it is important for AI scientists and engineers to have an understanding in the linguistic properties of their training and testing data. We hope that our published dataset and code contributions provide a foundation for future efforts towards Wu Chinese machine translation and language modeling.

\label{sec:bibtex}
\bibliography{anthology,custom}

\begin{thebibliography}{14}
\expandafter\ifx\csname natexlab\endcsname\relax\def\natexlab#1{#1}\fi

\bibitem[{Chen and Gussenhoven(2015)}]{Chen:15}
Yiya Chen and Carlos Gussenhoven. 2015.
\newblock \href {https://doi.org/10.1017/S0025100315000043} {Shanghai chinese}.
\newblock \emph{Journal of the International Phonetic Association}, 45(3):321–337.

\bibitem[{Chen(2023)}]{Chen:23}
Yuanhao Chen. 2023.
\newblock \href {http://arxiv.org/abs/2307.16199} {Improving tts for shanghainese: Addressing tone sandhi via word segmentation}.

\bibitem[{Eberhard et~al.(2024)Eberhard, Simons, and Fennig}]{Eberhard:24}
David~M. Eberhard, Gary~F. Simons, and Charles~D. Fennig. 2024.
\newblock \href {http://www.ethnologue.com} {Ethnologue: Languages of the world. twenty-seventh edition. dallas, texas: Sil international}.

\bibitem[{Joulin et~al.(2016)Joulin, Grave, Bojanowski, and Mikolov}]{joulin:16}
Armand Joulin, Edouard Grave, Piotr Bojanowski, and Tomas Mikolov. 2016.
\newblock Bag of tricks for efficient text classification.
\newblock \emph{arXiv preprint arXiv:1607.01759}.

\bibitem[{Lee et~al.(2022)Lee, Chen, Lam, Lau, and Tsui}]{Lee:22}
Jackson~L. Lee, Litong Chen, Charles Lam, Chaak~Ming Lau, and Tsz-Him Tsui. 2022.
\newblock Pycantonese: Cantonese linguistics and nlp in python.
\newblock \emph{Proceedings of the 13th Language Resources and Evaluation Conference}.

\bibitem[{{NLLB Team} et~al.(2022){NLLB Team}, Costa-jussà, Cross, Çelebi, Elbayad, Heafield, Heffernan, Kalbassi, Lam, Licht, Maillard, Sun, Wang, Wenzek, Youngblood, Akula, Barrault, Mejia-Gonzalez, Hansanti, Hoffman, Jarrett, Sadagopan, Rowe, Spruit, Tran, Andrews, Ayan, Bhosale, Edunov, Fan, Gao, Goswami, Guzmán, Koehn, Mourachko, Ropers, Saleem, Schwenk, and Wang}]{nllb-22}
{NLLB Team}, Marta~R. Costa-jussà, James Cross, Onur Çelebi, Maha Elbayad, Kenneth Heafield, Kevin Heffernan, Elahe Kalbassi, Janice Lam, Daniel Licht, Jean Maillard, Anna Sun, Skyler Wang, Guillaume Wenzek, Al~Youngblood, Bapi Akula, Loic Barrault, Gabriel Mejia-Gonzalez, Prangthip Hansanti, John Hoffman, Semarley Jarrett, Kaushik~Ram Sadagopan, Dirk Rowe, Shannon Spruit, Chau Tran, Pierre Andrews, Necip~Fazil Ayan, Shruti Bhosale, Sergey Edunov, Angela Fan, Cynthia Gao, Vedanuj Goswami, Francisco Guzmán, Philipp Koehn, Alexandre Mourachko, Christophe Ropers, Safiyyah Saleem, Holger Schwenk, and Jeff Wang. 2022.
\newblock \href {http://arxiv.org/abs/arXiv:1902.01382} {No language left behind: Scaling human-centered machine translation}.

\bibitem[{Norman(1988)}]{Norman:88}
Jerry Norman. 1988.
\newblock \emph{Chinese}.
\newblock Cambridge University Press.

\bibitem[{Pan et~al.(1991)Pan, Zhengzhang, You, and Chinfa}]{Pan:91}
Wuyun Pan, S.F. Zhengzhang, R.J. You, and Lien Chinfa. 1991.
\newblock \href {http://www.jstor.org/stable/23827040} {An introduction to the wu dialects}.
\newblock \emph{Journal of Chinese Linguistics Monograph Series}, (3):235--291.

\bibitem[{张惠英(2009)}]{Zhang:09}
张惠英. 2009.
\newblock \href {https://books.google.com.hk/books?id=TJXVSAAACAAJ} {\emph{崇明方言研究}}.
\newblock 中国社会科学出版社.

\bibitem[{张惠英 et~al.(2014)张惠英, 顾晓东, and 王洪钟}]{Zhang:14}
张惠英, 顾晓东, and 王洪钟. 2014.
\newblock \href {https://books.google.com.hk/books?id=4x4FoQEACAAJ} {\emph{崇明方言大词典}}.
\newblock 上海辞书出版社.

\bibitem[{赵元任(1956)}]{Zhao:56}
赵元任. 1956.
\newblock \href {https://books.google.com.hk/books?id=AbyDnQEACAAJ} {\emph{现代吴语的研究}}.
\newblock 科学出版社.

\bibitem[{钱乃荣(1992)}]{Qian:92}
钱乃荣. 1992.
\newblock \emph{当代吴语研究}.
\newblock 上海教育出版社.

\bibitem[{钱乃荣(2008)}]{Qian:08}
钱乃荣, editor. 2008.
\newblock \href {https://books.google.com.hk/books?id=gmiJQwAACAAJ} {\emph{上海话大词典}}.
\newblock 上海辞书出版社.

\bibitem[{闵家骥 et~al.(1986)闵家骥, 范晓, 朱川, and 张蒿岳}]{Min:86}
闵家骥, 范晓, 朱川, and 张蒿岳, editors. 1986.
\newblock \href {https://books.google.com.hk/books?id=0LgPAAAAYAAJ} {\emph{简明吴方言词典}}.
\newblock 上海辞书出版社.

\end{thebibliography}
\bibliographystyle{acl_natbib}




\end{CJK*}
\end{document}